\documentclass[a4paper]{svproc}
\usepackage{url}
\usepackage{graphicx}
\usepackage{amsmath} 
\usepackage{amssymb}  
\usepackage{bm}
\usepackage{color}
\usepackage{caption}

\begin{document}
\mainmatter              
\title{Online Adaptation for Flying Quadrotors in Tight Formations}
\titlerunning{Online Adaptation for Flying Quadrotors in Tight Formations}  
%
\author{Pei-An Hsieh\inst{1} \and Kong Yao Chee\inst{2} \and M. Ani Hsieh\inst{1}}
\authorrunning{Pei-An Hsieh et al.} 
%
\tocauthor{Pei-An Hsieh, Kong Yao Chee, and M. Ani Hsieh}
\institute{GRASP Laboratory,University of Pennsylvania, Philadelphia, PA 19104, USA,\\
\email{{pahsieh, mya}@seas.upenn.edu},
\and
DSO National Laboratories, Singapore, 12 Science Park Drive, Singapore 118225,\\
\email{ckongyao@alumni.upenn.edu}
}

\maketitle              

\begin{abstract}
The task of flying in tight formations is challenging for teams of quadrotors because the complex aerodynamic wake interactions can destabilize individual team members as well as the team.  Furthermore, these aerodynamic effects are highly nonlinear and fast-paced, making them difficult to model and predict. To overcome these challenges, we present ${\cal L}_1$ KNODE-DW MPC, an adaptive, mixed expert learning based control framework that allows individual quadrotors to accurately track trajectories while adapting to time-varying aerodynamic interactions during formation flights. We evaluate ${\cal L}_1$ KNODE-DW MPC in two different three-quadrotor formations and show that it outperforms several MPC baselines. Our results show that the proposed framework is capable of enabling the three-quadrotor team to remain vertically aligned in close proximity throughout the flight. These findings show that the ${\cal L}_1$ adaptive module compensates for unmodeled disturbances most effectively when paired with an accurate dynamics model.
\end{abstract}
\vspace{-2em}
\section{Introduction}
\vspace{-0.0em}
The ability to fly teams of quadrotors in tight formations is of significant interest in many commercial \cite{xu2025airbender}, scientific \cite{floreano2015science}, and military applications \cite{xiaoning2020analysis}. The key challenge in tight formation flights is the need to deal with the aerodynamic forces resulting from the turbulent wake interactions of the various propellers in a team, {\it e.g.}, downwash \cite{kiran2024downwash}. Downwash from a single quadrotor forms a turbulent jet that is highly nonlinear and difficult to model. When multiple quadrotors are involved, the resulting aggregated downwash becomes even more complex. Failure to account for these effects can lead to catastrophic collisions with other quadrotors in the formation or with the surrounding environment.

One strategy is to leverage the expressive power of deep neural networks to obtain an accurate model of downwash in multi-quadrotor scenarios \cite{shi2022neural,gielis2023modeling}. While these approaches are capable of capturing nontrivial aerodynamic interactions between nearby quadrotors, they often require a cumbersome data collection process or customized force sensors to obtain close-proximity interaction data. This poses a challenge for tight formation flights, as flying multiple quadrotors in close proximity without specialized controllers is infeasible. Moreover, the learned interactions may fail to generalize to formations with different configurations or numbers of quadrotors than those used during training \cite{gielis2023modeling}.

Recent advances in dynamics modeling have shown that mixed expert approaches can be used to develop highly predictive models for precise control with excellent sample-efficiency and generalizability \cite{chee2024flying,li2024learning}. Mixed expert models combine neural networks with first-principles models and have been shown to generate highly accurate models for various complex systems \cite{jiahao2021knowledge}.  These models have also been used in learning-based model predictive control (MPC) to achieve accurate autonomous driving \cite{li2024learning} and flying a pair of quadrotors in extremely tight formations \cite{chee2024flying}. However, a key limitation of these mixed expert models is that they are often trained offline and cannot adapt to unseen conditions during deployment. This limits their effectiveness in applications such as mid-air battery replacement \cite{jain2020flying}, flying through confined urban canyons \cite{folk2024learning}, and cooperative aerial manipulation tasks \cite{xu2025airbender} where vehicles may experience more complex aerodynamic interactions with other vehicles and/or the environment.

To address time-varying and difficult-to-model uncertainties, adaptive controllers like $\mathcal{L}_1$ provide a computationally and sample efficient way to achieve responsive online adaptation \cite{wu2024mathcall1quadmathcall1adaptiveaugmentation,hanover2021performance}.  Incorporating unmodeled force and torque estimates from an $\mathcal{L}_1$ adaptive module into MPC has been shown to stabilize quadrotors subjected to significant unmodeled or out-of-distribution disturbances \cite{chee2023enhancing}. It is important to note that in these works, no additional modeling of disturbances was included, so it remains unclear whether $\mathcal{L}_1$ adaptive control can benefit from more accurate dynamic models of the plant and/or disturbances.

In this work, we present $\mathcal{L}_1$ KNODE-DW MPC, which incorporates an $\mathcal{L}_1$ adaptive module into our existing mixed expert KNODE-DW MPC framework \cite{chee2024flying}. KNODE-DW MPC is a learning-based MPC framework that combines a first-principles physics-based model with Neural ODEs to accurately model and compensate for downwash dynamics. The objective is to investigate whether accurate modeling of disturbances enhances the effectiveness of the $\mathcal{L}_1$ adaptive module in enabling individual quadrotors flying in tight formations to adapt to unmodeled aerodynamic interactions.
\vspace{-2em}
\paragraph{\bf Contribution} We present the ${\cal L}_1$ KNODE-DW MPC framework to enable close proximity flight for a team of quadrotors. This framework leverages the complementary strengths of mixed expert models and adaptive control to compensate for large and time-varying disturbances. Through extensive physical experiments, we demonstrate that ${\cal L}_1$ KNODE-DW MPC outperforms several baseline MPC strategies.  Our results show the importance of combining an adaptive control module with an accurate dynamics model to yield the best performance. We validated our strategy using a team of three quadrotors tasked to fly in a vertically aligned formation while maintaining a pair-wise separation distance of two body lengths.  To the best of our knowledge, this is the first demonstration of such a compact formation involving more than two quadrotors.

\section{Technical Approach}\label{sec:techapp}
\vspace{-0.5em}
The objective of this work is to develop a control framework to enable teams of quadrotors to fly in tight formations subject to complex aerodynamic interactions.  To accomplish this, we build upon the strategies developed in \cite{chee2024flying} and \cite{chee2023enhancing} and describe our proposed framework below.  We refer the interested reader to \cite{chee2024flying,chee2023enhancing} for the relevant details.
\vspace{-1em}
\vspace{-0.5em}
\subsection{Modeling Complex Aerodynamic Interactions}
\vspace{-0.5em}
To model the aerodynamic interactions resulting from formation flight in a team of quadrotors, we extend the KNODE-DW MPC and DW MPC framework described in \cite{chee2024flying}. Given a team of quadrotors, let $i$ denote the $i^{th}$ quadrotor and $p_i = [p_{x_i} \, p_{y_i} \, p_{z_i}]^{\top}$ denote the quadrotor's position in $\mathbb{R}^3$ where $p_{z_i}$ corresponds to the vehicle's vertical height. We denote each quadrotor's state as $x_i = [p_i^{\top}, v_i^{\top}, q_i^{\top}]^{\top} \in \mathbb{R}^{10}$, where $v_i \in \mathbb{R}^3$ is the velocity and $q_i = [q_x, q_y,q_z,q_w]^{\top} \in \mathbb{R}^4$ is the quaternion. Let ${\cal N}_i(p_i) = \{p \mid \Vert p-p_i \Vert \leq \alpha\}$ denote the neighborhood of $i$ with $\alpha > 0$ denoting the range in which aerodynamic interactions between $i$ and any $m \in {\cal N}_i$ are non-negligible. The set ${\cal S}_i = \{x_m \mid m \in {\cal N}_i \}$ represents the collection of states of neighboring quadrotors in ${\cal N}_i$. Based on existing empirical downwash force measurements \cite{gielis2023modeling}, we assume that for all $m \in {\cal N}_i$, the aggregated force of on $i$ can be approximated as a pairwise summation of the force from each $m \in {\cal N}_i$. As such, we model the total force experienced by $i$ exerted by all $m \in {\cal N}_i$ such that $p_{z_m} > p_{z_i}$ as
\vspace{-0.5em}
\begin{equation}
{\cal T}_{d,i}(x_i, {\cal S}_i) = \sum_{m \in {\cal N}_i,\, p_{z_m} > p_{z_i}} \tau_{d,m}(x_i, x_m),
\vspace{-0.5em}
\end{equation}
where $\tau_{d,m}$ is the force exerted by $m$ on $i$. In general, $\tau_{d,m}$ can be modeled solely by a first-principles physics-based model, a purely data-driven one, or a combination of both. 

In this work, we consider both a first-principles physics-based model for $\tau_{d,m}$, denoted as DW, and a hybrid first-principles and data-driven model, referred to as a mixed expert model denoted as KNODE-DW. The DW model is derived from airflow velocity and drag equations, and KNODE-DW combines DW and Neural ODEs to improve modeling accuracy. Both models are described in detail in \cite{chee2024flying}. To train the data-driven component of our KNODE-DW model, we use the same training data as in \cite{chee2024flying}, which was collected from two quadrotors flying in the {\it static top} and {\it stacked} formations.  The {\it static top} formation consists of a quadrotor hovering above and another quadrotor flying a straight trajectory beneath it, while the {\it stacked} formation consists of two vertically aligned quadrotors flying in the same direction. 
\vspace{-1.2em}
\subsection{Formation Controller Synthesis}
\vspace{-0.5em}
In this work, we integrate an $\mathcal{L}_1$ adaptive module with downwash models of varying fidelity to examine improvements in trajectory tracking performance for a team of quadrotors. We select the control inputs as $u:=[u_{\gamma},\omega^{\top}]^{\top}\in\mathbb{R}^4$ where the commanded thrust is given by $u_{\gamma}$ and the commanded angular rates are denoted as $\omega \in \mathbb{R}^3$. Starting with {\small $f_{nom}(x,u)$}, the nominal quadrotor dynamics model is defined as
\vspace{\baselineskip}
\begin{equation}\label{eq:eoms}
\begin{split}
    \dot{p} = v,\quad
    \dot{v} = -g + (1/\eta)R \gamma, \quad
    \dot{q} = \frac{1}{2} G(q)^\top \omega,
\end{split}
\end{equation}
where $g\in\mathbb{R}^3$ is the gravity vector in world frame and $R\in \mathbb{R}^{3\times3}$ denotes the rotation matrix from the body frame to the world frame.  The mass of the quadrotor is denoted as $\eta$ and the thrust vector is given by $\gamma = [0,0,u_{\gamma}]^{\top} \in \mathbb{R}^3$. The matrix $G(q)$ that maps angular rates to the quaternion derivatives is defined as
\vspace{-0.5em}
\[
G(q) =
\begin{bmatrix}
  q_w & q_z & -q_y & -q_x \\
  -q_z &  q_w & q_x & -q_y \\
   q_y &  -q_x & q_w & -q_z 
\end{bmatrix}.
\vspace{-0.5em}
\]
We employ a standard MPC strategy to both highlight the challenging nature of the task and to serve as a baseline. We then compare the nominal MPC strategy with the physics-only (DW) and mixed expert (KNODE-DW) MPC strategies.  Lastly, we incorporate an ${\cal L}_1$ adaptive module to all three frameworks which we will refer to as $\mathcal{L}_1$ MPC, $\mathcal{L}_1$ DW MPC, and $\mathcal{L}_1$ KNODE-DW MPC. 
\vspace{-1em}
\paragraph{\bf Baseline MPC} 
The optimal control problem of the nominal MPC for quadrotor $i$ over the interval $t \in [kT, (k+1)T]$ is formulated as
\begin{subequations} \label{eq:dwmpc}
\vspace{-0.23cm}
\begin{align}
    \underset{\{x_{i,j}\},\{u_{i,j}\}}{\textnormal{min.}}\quad\; & \sum_{j=0}^{N-1} \left|\left|x_{i,j}- x_{i,j}^{ref}(k)\right|\right|^2_Q + ||u_{i,j}||_R^2 + \left|\left|x_{i,N}-x_{i,N}^{ref}(k)\right|\right|^2_P \label{eq:term_mpc_cost}\\
    \text{s.t.}\quad\; &x_{i,j+1} = f_{nom}(x_{i,j}, u_{i,j}), \label{eq:dynamics_const}\\ 
    \; &x_{i,j} \in \mathcal{X}_i, \quad u_{i,j} \in \mathcal{U}_i, \quad j=0,\dots,N-1,\\
    \; &x_{i,N} \in \mathcal{X}_{i}^{f},\quad x_{i,0} = x_i(k), \label{eq:term_constraint}
\end{align}
\end{subequations} 
where $T$ is the sampling period and $k\in \mathbb{N}$. The prediction horizon is denoted as $N \in \mathbb{N}_+$ and $x_i(k)$ denotes the state measurement at timestep $k$. The notation $||\delta||_A$ stands for $\delta^{\top} A \delta$ for vector $\delta$ and matrix $A$. 
The matrices $Q$, $R$ and $P$ penalize deviation from the reference states, control inputs, and terminal states. The corresponding constraints are defined by sets $\mathcal{X}_i$, $\mathcal{U}_i$, and $\mathcal{X}_i^f$. A new solution to \ref{eq:dwmpc} is obtained at each time step and only the first element of the solution sequence, $u_{i,0}^{\star}$, is executed. 
\vspace{-1em}
\paragraph{\bf $\mathcal{L}_1$ Adaptive Module}
Inspired by $\mathcal{L}_1$ adaptive control, the adaptive module estimates and compensates for unmodeled disturbances online through a sequential procedure consisting of a state estimator, an adaptation law, and a low-pass filter, as described in \cite{chee2023enhancing}. The state predictor considers the partial states $z_i=[v_i^{\top}]^{\top} \in \mathbb{R}^3$, whose dynamics is given by
\vspace{-0.5em}
\begin{equation}
\dot{\hat{z}}_i = -g + (1/\eta){\cal T}_{d,i}(x_{i}, {\cal S}_i) + B\bar{u}_{\gamma,i} + \hat{\sigma}_i + A(\hat{z}_i-z_i). 
\end{equation}
The matrix $A \in \mathbb{R}^3$ is a user-selected Hurwitz matrix, and $\bar{u}_{\gamma,i}$ denotes the commanded thrust from the previous timestep. The matrix $B := [(1/\eta)Re_3]$ maps the commanded thrust to the resulting acceleration, where $e_3\in\mathbb{R}^3$ represents the $z$-axis of the body frame. For {\small$t \in [kT, (k+1)T]$}, the vector $\hat{\sigma}_i \in \mathbb{R}^3$ denotes the residual acceleration, which is estimated using the following piecewise constant adaptation law,
\vspace{\baselineskip}
\begin{equation}
    \hat{\sigma}(t) := \hat{\sigma}(kT) := (e^{AT}-\mathbb{I})^{-1}Ae^{AT}(z_i-\hat{z}_i).
\end{equation}
The residual accelerations are filtered using a low-pass filter and then incorporated into the MPC as unmodeled dynamics for quadrotor $i$. The resulting term, denoted $U_{\sigma,i}(t)$ for $t \in [kT, (k+1)T]$, is given by
\vspace{-0.5em}
\begin{equation}
U_{\sigma,i}(t) := U_{\sigma,i}(kT):= (U_{\sigma,i}((k-1)T) + \hat{\sigma}(kT))e^{-\alpha T} - \hat{\sigma}(kT),
\vspace{-0.5em}
\end{equation}
where $\alpha$ is the cut-off frequency of the low-pass filter. 
\vspace{-1em}
\paragraph{\bf $\mathcal{L}_1$ KNODE-DW MPC}
We incorporate the downwash and estimated unmodeled dynamics into the MPC to accurately track trajectories during tight formation flights. This is done by modifying the model in \eqref{eq:dynamics_const} to 
\vspace{-0.5em}
\begin{equation}
x_{i,j+1} := {f_{nom}(x_{i,j}, u_{i,j}) + f_{d,i}(x_{i,j}, {\cal S}_i) + f_{\sigma,i}(x_i(k), \bar{u}_{\gamma,i}, {\cal S}_i,t)},
\end{equation}
where $f_{d,i} = [0_{1\times 3}, (1/\eta){\cal T}_{d,i}^{\top}, 0_{1\times 4}]^{\top} \in \mathbb{R}^{10}$ encodes the downwash forces, and $f_{\sigma,i} = [0_{1\times 3}, U_{\sigma,i}^{\top}, 0_{1\times 4}]^{\top} \in \mathbb{R}^{10}$ accounts for the estimated unmodeled dynamics. The key difference among $\mathcal{L}_1$ KNODE-DW MPC, $\mathcal{L}_1$ DW MPC, and $\mathcal{L}1$ MPC lies in how ${\cal T}_{d,i}$ is constructed, with mixed expert models in $\mathcal{L}_1$ KNODE-DW MPC, physics-based models in $\mathcal{L}_1$ DW MPC, and omitted entirely in $\mathcal{L}_1$ MPC. Their counterparts without the $\mathcal{L}_1$ adaptation module are formed by removing the term $f_{\sigma,i}(x_i(k), \bar{u}_{\gamma,i}, {\cal S}_i, t)$ from the model, resulting in
\vspace{-0.5em}
\begin{equation}
x_{i,j+1} := {f_{nom}(x_{i,j}, u_{i,j}) + f_{d,i}(x_{i,j}, {\cal S}_i)}.
\end{equation}

\vspace{-2.5em}
\section{Results}
\vspace{-1.0em}
We conducted physical experiments using three Crazyflie 2.1 quadrotors. Each Crazyflie has a diameter of 0.1 m and weighs approximately 34 g. They are equipped with the thrust upgrade bundles from Bitcraze to improve their control authority. Pose estimates for each quadrotor are provided using a Vicon motion capture system at 120 Hz.  The control frameworks were implemented using \texttt{acados} \cite{Verschueren2021}, and the control inputs are thrust and angular rates of the vehicle. The control inputs for the bottom quadrotor in the formation were computed using an Intel NUC13 running on an Intel i7 CPU, which communicates with the Crazyflie via a Crazyradio PA at 200 Hz. The control inputs for the other two quadrotors were computed using a laptop with an Intel i5 CPU, communicating with the Crazyflies through another Crazyradio PA at 400 Hz.
\noindent
\begin{minipage}{\textwidth}
    \centering
    \includegraphics[scale=0.26, trim = -0.2cm 0.5cm 0cm 0.0cm, clip]{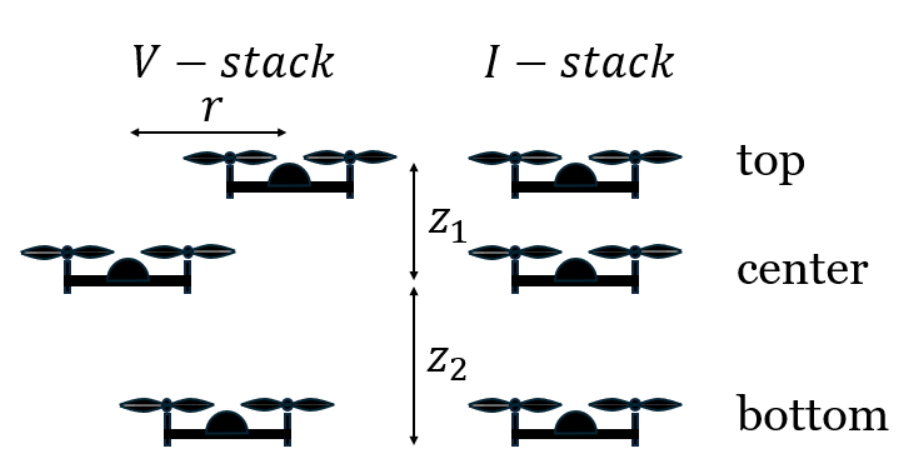}
    \captionof{figure}{\small \textbf{Schematics of \textit{V-stack} and \textit{I-stack} formations:} $z_1$ and $z_2$ represent the vertical separations between the top and center quadrotors, and between the center and bottom quadrotors, respectively. \textit{r} denotes the horizontal separation of the top and the center quadrotors in \textit{V-stack}.}
    \label{fig:formations}
\end{minipage}

To evaluate the efficacy of control frameworks, we considered two different stacked formations as shown in Fig. \ref{fig:formations}. The first formation, \textit{V-stack}, involves quadrotors that are horizontally displaced along a vertical centerline but still positioned within each other’s downwash regions. The horizontal position of the bottommost quadrotor in \textit{V-stack} is assigned to be at the center of the displacement. The second formation consists of three vertically aligned quadrotors, which we denote as \textit{I-stack}. These configurations were chosen to test the control frameworks' ability to compensate for both transient and steady-state downwash forces during formation flight. We will refer to the topmost, middle, and bottommost quadrotors in the formation as the top, center, and bottom quadrotors respectively. Each test scenario is conducted five times, and two trajectory tracking metrics were evaluated: the root mean square error (RMSE) and the maximum deviation in the z-direction ($z_{max}$).

The control frameworks were first tested on the center quadrotor and then on the bottom quadrotor.  Based on the results, we then employed the best performing strategies to fly the team in even tighter formations.  

\noindent
\begin{minipage}{\textwidth}
    \centering
    \includegraphics[width=0.85\textwidth, trim=0.2cm 0.5cm 0cm 1.5cm, clip]{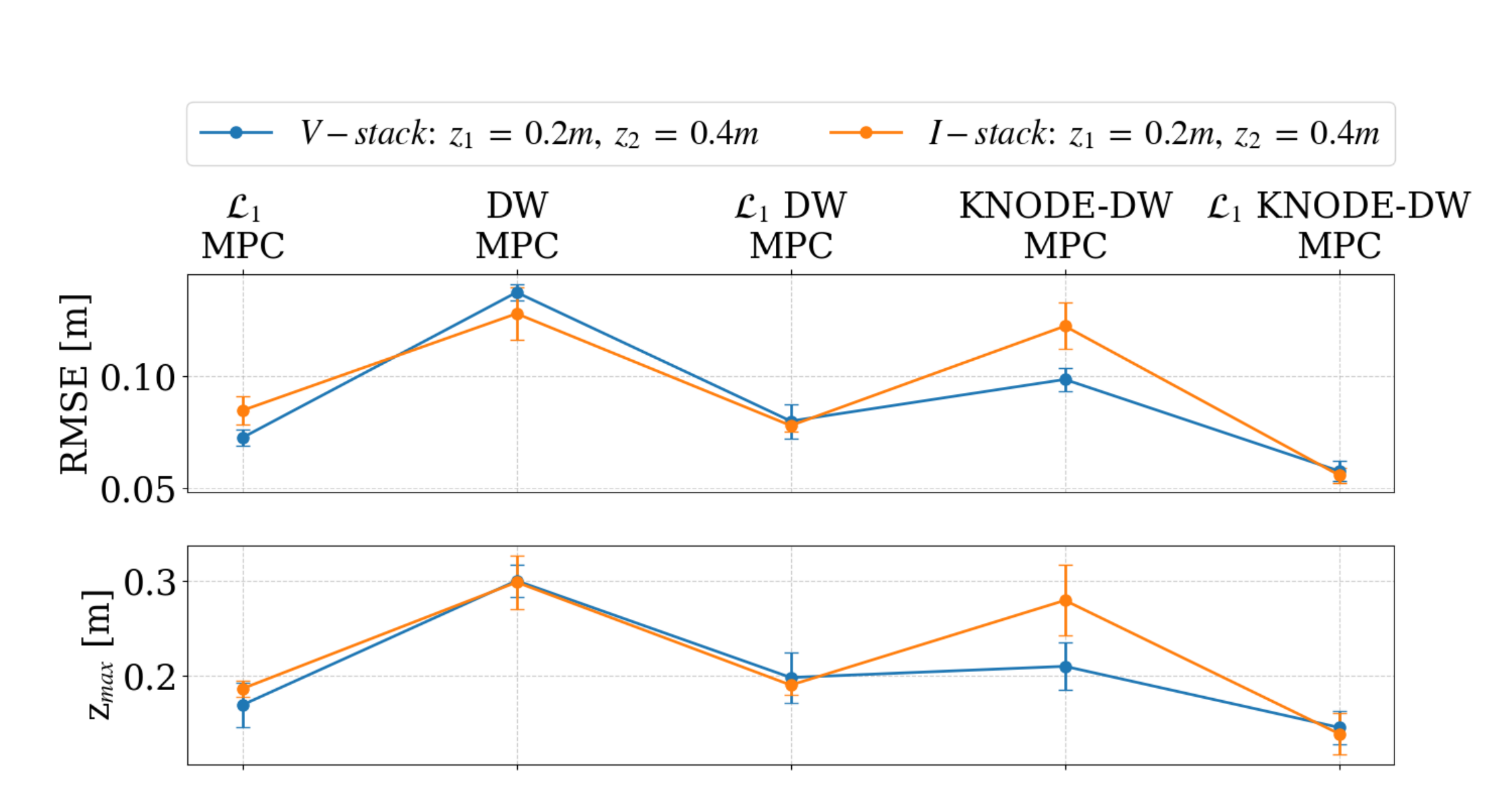}
    \captionof{figure}{\small\textbf{Center quadrotor statistics:} Performance of the center quadrotor flying in \textit{V-stack} and \textit{I-stack} formations. The top subplot shows RMSEs, and the bottom subplot shows the maximum vertical deviation $z_{\max}$. Markers and error bars represent the mean and standard deviation.}
    \label{fig:mid_quad}
\end{minipage}
\vspace{-1em}
\paragraph{\bf Performance of the Center Quadrotor}\label{sec:exp_mid}
For the center quadrotor, we implemented and validated all the control frameworks described in in Section \ref{sec:techapp} with the exception of the nominal MPC.  This is because the nominal MPC strategy lead to collisions between the center and bottom quadrotors.  Since the tracking performance of a vehicle is impacted by the performance of its neighbors, we implemented the best performing control architectures in the neighboring quadrotors based on results reported in previous works \cite{chee2024flying,chee2023enhancing}. As such, the $\mathcal{L}_1$ MPC and KNODE-DW MPC  were implemented on the top and bottom quadrotors for all test cases to ensure controlled experimental conditions. The \textit{V-stack} formation uses $r = 0.1$ m, $z_1 = 0.2$ m, and $z_2 = 0.4$ m. The \textit{I-stack} formation also uses $z_1 = 0.2$ m, and $z_2 = 0.4$ m. 

As depicted in Fig. \ref{fig:mid_quad}, the proposed $\mathcal{L}_1$ KNODE-DW MPC provides the best performance in both \textit{V-stack} and \textit{I-stack} formations for the center quadrotor. On average, $\mathcal{L}_1$ KNODE-DW MPC improved RMSE by 47.9 \% and $z_{max}$ by 40.4 \% compared to KNODE-DW MPC. Incorporating the $\mathcal{L}_1$ adaptive module into all the MPC frameworks significantly improved trajectory tracking performance compared to their counterparts without the module. Notably, the $\mathcal{L}_1$ MPC also achieved a 28.5 \% improvement in RMSE and a 26.3 \% improvement in $z_{max}$ compared to KNODE-DW MPC. Our results demonstrates the effectiveness of $\mathcal{L}_1$ adaptive module in compensating for unmodeled dynamics.  
\vspace{-0em}



\noindent
\begin{minipage}{\textwidth}
    \centering
    \includegraphics[width=0.85\textwidth, trim=0.2cm 0.2cm 0cm 0.5cm, clip]{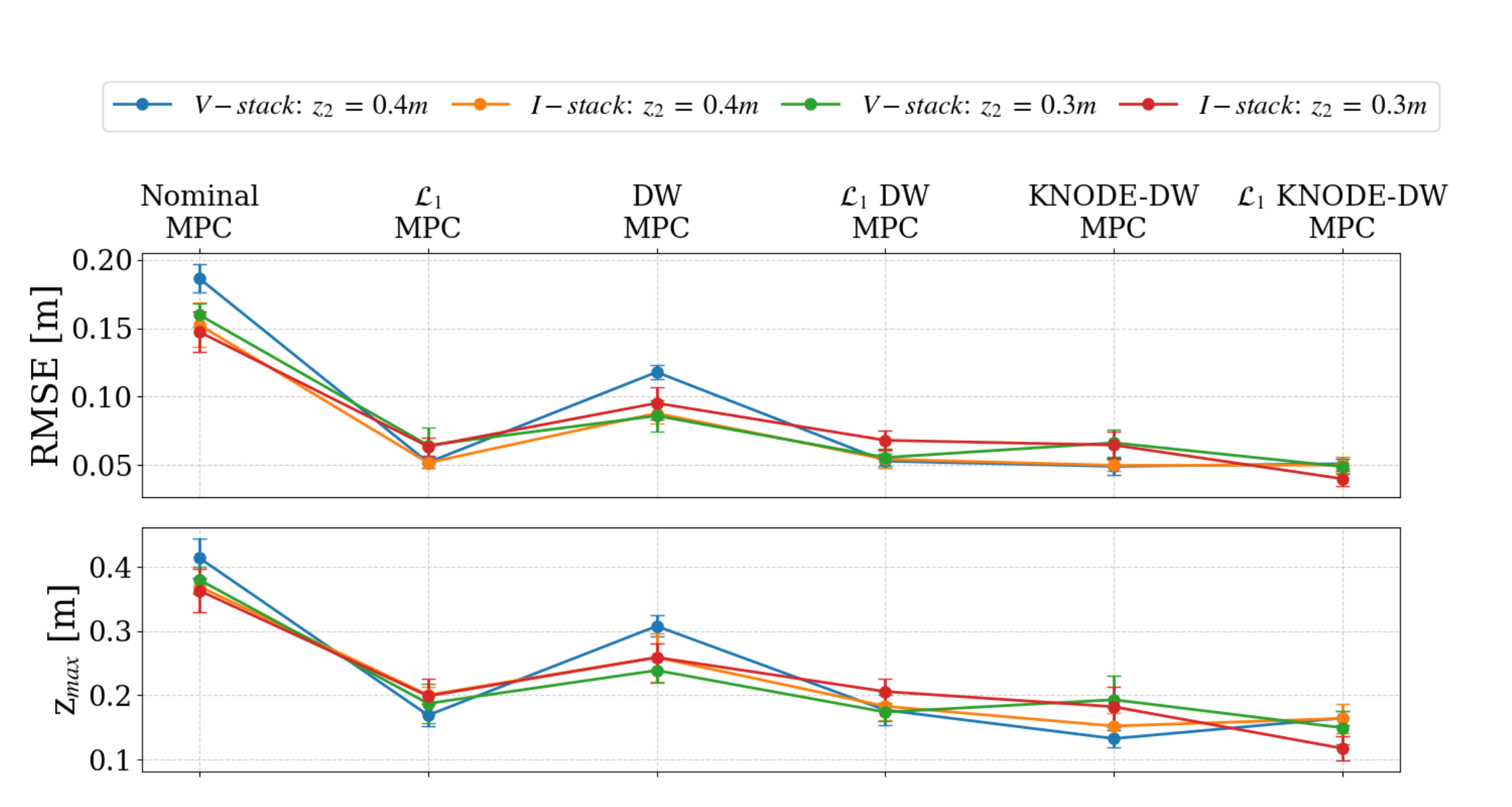}
    \captionof{figure}{\small\textbf{Bottom quadrotor statistics:} Performance of the bottom quadrotor in \textit{V-stack} and \textit{I-stack} formations. The top subplot shows RMSEs, and the bottom subplot shows $z_{\max}$. Markers and error bars represent the mean and standard deviation.}
    \label{fig:bottom}
\end{minipage}
\vspace{-1.0em}
\paragraph{\bf Performance of the Bottom Quadrotor}\label{sec:exp_bot}
Similarly, we implemented and validated all the control frameworks described in Section \ref{sec:techapp} on the bottom quadrotor. Given the strong performance of $\mathcal{L}_1$ KNODE-DW MPC in the experiment for the middle quadrotor, we used $\mathcal{L}_1$ KNODE-DW MPC on the top and center quadrotors. We tested two vertical separations between the middle and bottommost quadrotors:  $z_2 = 0.3$ m and $z_2 = 0.4$ m. The \textit{V-stack} formation uses $r = 0.1$ m, $z_1 = 0.2$ m, and the \textit{I-stack} formation also uses $z_1 = 0.2$ m.  

\begin{figure}
\centering

\begin{minipage}{0.49\textwidth}
    \centering
    \includegraphics[width=\linewidth, trim=1.0cm 0.5cm 0.0cm 0.0cm, clip]{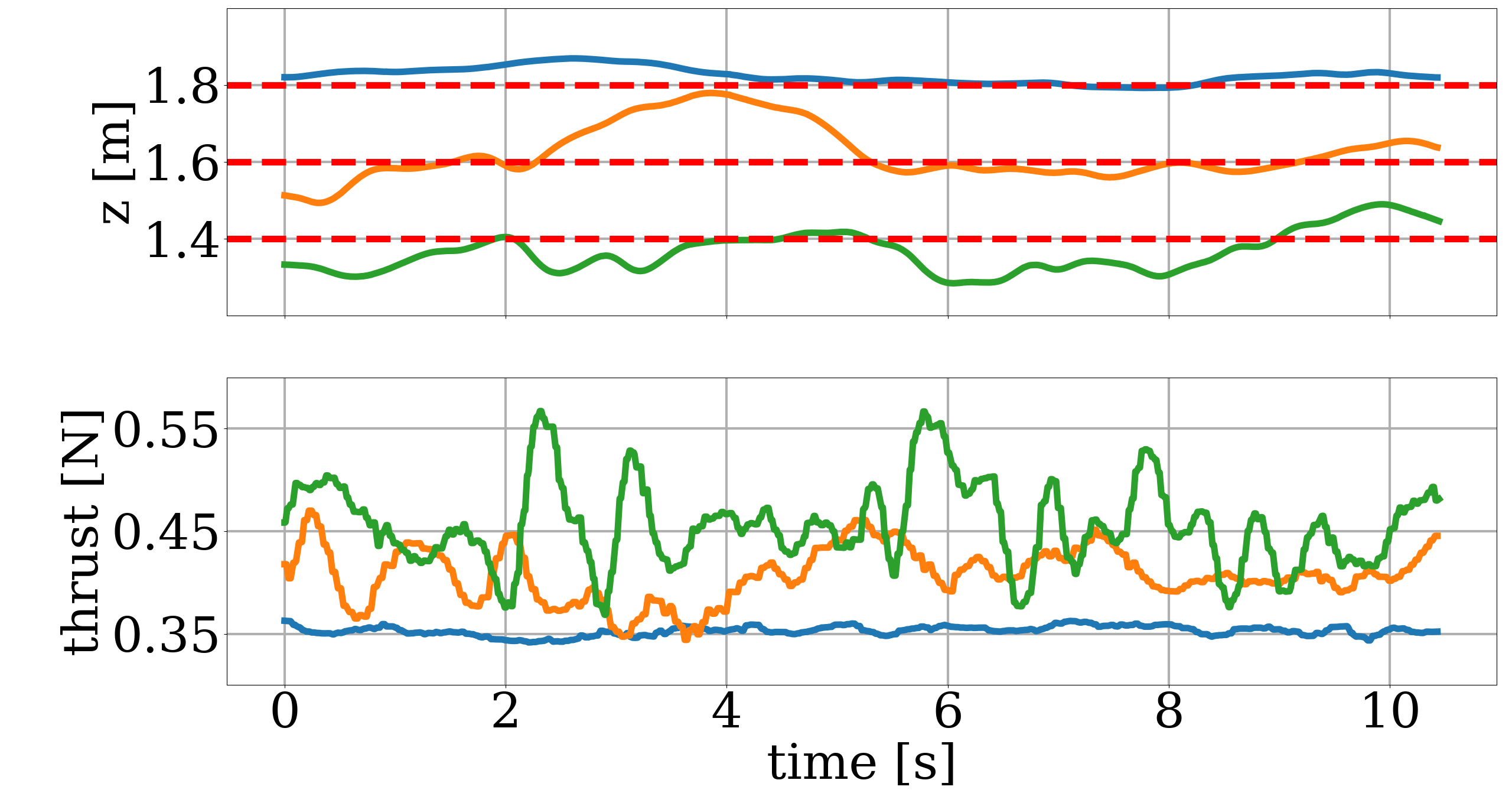}
\end{minipage}
\hfill
\begin{minipage}{0.49\textwidth}
    \centering
    \includegraphics[width=\linewidth, trim=3.0cm 2.0cm 5.0cm 1.0cm, clip]{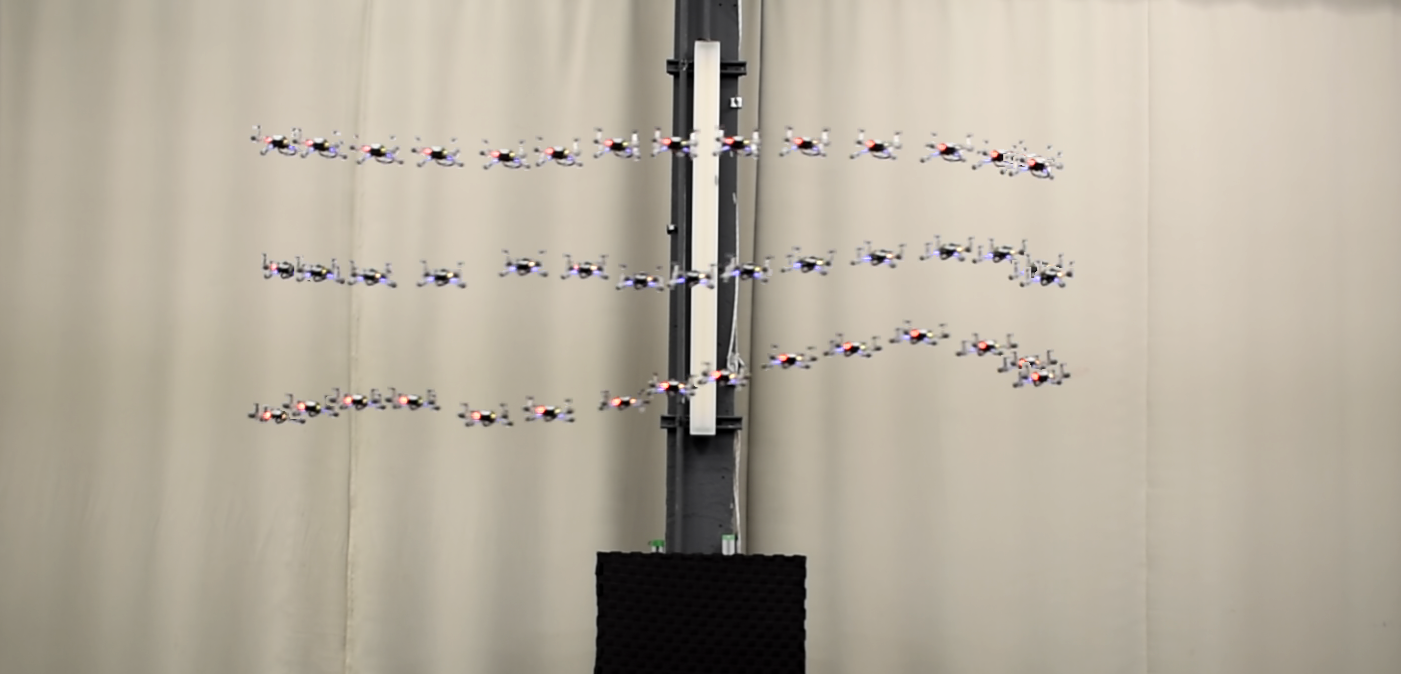}
\end{minipage}

\caption{\small\textbf{Time history and real-world trajectory:} The plot on the left shows the $z$ position and thrust histories of the three quadrotors flying a tight \textit{I-stack} formation. The blue, orange, and green lines represent the top, center, and bottom quadrotors, respectively. The composite photo on the right captures the second half of the same real-world experiment, showing the quadrotors in flight.}
\label{fig:time_hist_plus_photo}
\end{figure}

The results of these four configurations are shown in Fig. \ref{fig:bottom}. We note that all methods significantly improve trajectory tracking performance compared to the nominal MPC. $\mathcal{L}_1$ KNODE-DW MPC performs the best overall in both RMSE and $z_{max}$, reducing the RMSE by 15 \% and $z_{max}$ by 6.6 \% over KNODE-DW MPC. It is worth noting that KNODE-DW MPC, $\mathcal{L}_1$ DW MPC, and $\mathcal{L}_1$ MPC were able to match the performance of $\mathcal{L}_1$ KNODE-DW MPC in formations with $z_2 = 0.4$ m. However, in formations with a smaller $z_2 = 0.3$ m where the aggregate downwash is stronger, $\mathcal{L}_1$ KNODE-DW MPC was the only framework to maintain similar tracking performances as $z_2 = 0.4$ m.  
\vspace{-1em}
\paragraph{\bf Achieving Tight \textit{I-stack} Formations}\label{sec:ext_tight}
From the results shown in Fig. \ref{fig:mid_quad} and \ref{fig:bottom}, the best-performing framework for both the middle and bottom quadrotors is $\mathcal{L}_1$ KNODE-DW MPC. We then commanded the team to fly in an \textit{I-stack} formation with $z_1 = 0.2$ m and $z_2 = 0.2$ m.  The time histories and a composite photo for the flight are shown in Fig. \ref{fig:time_hist_plus_photo}. The team successfully completed the formation flight while maintaining a vertical separation of less than six body lengths throughout the flight. The RMSEs for the top, center, and bottom quadrotors are $0.034$ m, $0.057$ m, and $0.049$ m, respectively, while the corresponding $z_{max}$ values are $0.069$ m, $0.179$ m, and $0.115$ m. 





\vspace{-1.2em}

\vspace{-1.0em}
\section{Discussions}
\vspace{-0.5em}
From our experiments, we see that aerodynamic interactions significantly impacts the tracking performance.
Previous works \cite{chee2024flying,gielis2023modeling} only consider the effects of downwash forces generated by one quadrotor flying above another one.  However, from our experiments we note that in more complex formations, individual vehicles are impacted by both the cummulative downwash forces generated from vehicles above and the aerodynamics created by vehicles below. This can be seen in our experiments where the center quadrotor's ability to track its own trajectory is significantly impacted by the bottom quadrotor's existence.  

This is supported by Fig. \ref{fig:mid_quad} and \ref{fig:bottom}, where all control frameworks improves the performance of the bottom quadrotor compared to the center quadrotor. For tighter formations as shown in Fig. \ref{fig:time_hist_plus_photo}, the middle quadrotor also exhibits a larger RMSE and $z_{max}$ relative to the reference trajectory. These large deviations were not observed in previous work where only two quadrotors in an equivalent  \textit{I-stacked} formation with a separation of $0.2$ m was considered \cite{chee2024flying}. As such, in larger formations, vehicles placed closer to the center of the formation will have to simultaneously contend with the cumulative downwash forces from the vehicles above it, as well as the aerodynamic interactions resulting from vehicles below it. This makes synthesizing robust controllers for vehicles in the center of large formations particularly challenging.


Our results also show how the $\mathcal{L}_1$ adaptive module improves with accurate modeling of disturbances.
$\mathcal{L}_1$ KNODE-DW MPC, $\mathcal{L}_1$ DW MPC, and $\mathcal{L}_1$ MPC all show improvement over their counterparts without the $\mathcal{L}_1$ adaptive module. This suggests that the $\mathcal{L}_1$ module can enhance MPC performance when the model within the MPC is imperfect. From the experiments in \cite{chee2024flying}, the downwash models can be ranked by accuracy as follows: the mixed expert KNODE-DW model is the most accurate, followed by the physics-based DW model, and lastly, no downwash model is the least accurate of all. This can be seen in the results presented in Fig. \ref{fig:bottom}. Furthermore, these results also suggest that the physics-based DW model is not particularly accurate, since the difference in performance between the $\mathcal{L}_1$ MPC and $\mathcal{L}_1$ DW MPC is quite small. 
On average, we observe that $\mathcal{L}_1$ KNODE-DW MPC performs the best, followed by $\mathcal{L}_1$ DW MPC, with $\mathcal{L}_1$ MPC performing the worst. This result indicates that the $\mathcal{L}_1$ adaptive module benefits from being paired with a more accurate disturbance model. 


Lastly, while the $\mathcal{L}_1$ KNODE-DW MPC framework performs the best, the $\mathcal{L}_1$ MPC framework shows strong performance.
$\mathcal{L}_1$ KNODE-DW MPC provides the best overall trajectory tracking performance, achieving the smallest RMSE and $z_{\max}$ with a low standard deviation. The tight \textit{I-stack} formation was only achievable with the center quadrotor using $\mathcal{L}_1$ KNODE-DW MPC, as other frameworks typically resulted in collisions with the bottom quadrotor. Despite being outperformed by $\mathcal{L}_1$ KNODE-DW MPC in all cases, $\mathcal{L}_1$ MPC was only slightly worse than KNODE-DW MPC and $\mathcal{L}_1$ DW MPC as seen in Fig. \ref{fig:mid_quad} and \ref{fig:bottom}. We believe this is because the $\mathcal{L}_1$ adaptation module is able to successfully compensate for the downwash force since it includes an online state prediction component. As such, the $\mathcal{L}_1$ adaptive control is a good alternative when computational efficiency is critical and tracking precision requirements are less stringent.

However, since the $\mathcal{L}_1$ adaptive module lacks a downwash model, it is not as responsive as $\mathcal{L}_1$ KNODE-DW MPC when entering or exiting downwash regions, as the latter has a highly predictive model of the downwash forces.  In addition, the absence of a downwash model sacrifices robustness in scenarios where the downwash is too strong to be compensated for. As such, a strategy that relies solely on the $\mathcal{L}_1$ adaptive controller may be too slow to respond to fast changing and/or very large disturbances.  
\vspace{-1.2em}
\section{Conclusion and Future Work}
\vspace{-0.5em}
This work presents $\mathcal{L}_1$ KNODE-DW MPC for quadrotor teams flying in tight formations. Our experiments show how the aerodynamic interactions between quadrotors significantly impact the team's ability to maintain tight formations. We demonstrated that $\mathcal{L}_1$ KNODE-DW MPC outperforms all other baselines for quadrotors in different positions within the formation and is crucial for maintaining small vertical separations. We also observed that the $\mathcal{L}_1$ adaptive module performs better when supported by a more accurate dynamics model. The mixed expert KNODE-DW model provides an accurate plant model for MPC, while the $\mathcal{L}_1$ adaptive module successfully compensates for inaccuracies between the mixed expert model and the true system dynamics. A potential future direction is to incorporate collision avoidance constraints in these tight formations, improving the reliability of our framework while preserving its performance for broader applications for formation flight.
\paragraph{\bf Acknowledgements} We gratefully acknowledge the support of ONR Awards N000142512171 and N000142512070.
%
%

\end{document}